\documentclass[journal]{IEEEtran}
\usepackage{comment}
\IEEEoverridecommandlockouts
\usepackage{multirow}

\usepackage[noadjust]{cite}

\usepackage{amsmath,amssymb,amsfonts}
\usepackage{algorithm}
\usepackage[noend]{algpseudocode}

\usepackage{graphicx}
\usepackage{amsmath}
\usepackage{textcomp}
\usepackage{mathtools}
\usepackage{xcolor}
\usepackage{float}

\usepackage[justification=centering]{caption}
\renewcommand{\baselinestretch}{0.93}
\def\BibTeX{{\rm B\kern-.05em{\sc i\kern-.025em b}\kern-.08em
    T\kern-.1667em\lower.7ex\hbox{E}\kern-.125emX}}

\usepackage{booktabs}

\ifCLASSINFOpdf
\else
\fi
\begin{document}
%
\title{Deep Generative Models for Vehicle Speed Trajectories}
%
%
%

\author{
    \IEEEauthorblockN{
        Farnaz Behnia\IEEEauthorrefmark{1}, 
        Dominik Karbowski\IEEEauthorrefmark{2}, 
        Vadim Sokolov\IEEEauthorrefmark{1}}
    \IEEEauthorblockA{
            \begin{center}
            \begin{tabular}{ c c }
                \IEEEauthorrefmark{1}Department of Systems Engineering and Operations Research &
                \IEEEauthorrefmark{2} Argonne National Laboratory, IL, U.S.A.\\
                George Mason University, Fairfax, VA, U.S.A. & 
                {e-mail: dkarbowski@anl.gov}\\
                e-mail: \{fbehnia,vsokolov\}@gmu.edu &
                {} \\
            \end{tabular}
            \end{center}
    } 
}

\maketitle

\begin{abstract}
Generating realistic vehicle speed trajectories is a crucial component in evaluating vehicle fuel economy and in predictive control of self-driving cars. Traditional generative models rely on Markov chain methods and can produce accurate synthetic trajectories but are subject to the curse of dimensionality. They  do not allow to include conditional input variables into the generation process. In this paper, we show how extensions to deep generative models allow accurate and scalable generation. Proposed architectures involve recurrent and feed-forward layers and are trained using adversarial techniques. Our models are shown to perform well on generating vehicle trajectories using a model trained on GPS data from Chicago metropolitan area.
\end{abstract}


%
\IEEEpeerreviewmaketitle

\section{Introduction}
%
%
%
%
\IEEEPARstart{S}{everal}  intelligent transport applications, such as  optimizing vehicle energy consumption \cite{opt-eng,Zhang17}, eco-routing \cite{eco,Wei11,Zeng18,Fei18,Grant14,Rezaei15}, driver assistance \cite{Liebner13,Jing15} and range predictions for electric vehicles \cite{rangep,Vatanparvar19,Scheubner14} rely on prediction and generation of vehicle speed trajectories. Our  work is motivated by the problem of evaluating energy consumption of a vehicle fleet, given the routes of each vehicle in the fleet. The goal is to generate a  ``library'' of realistic and representative trajectories that can be used as inputs to energy estimation models.

The problem of speed generation is defined as follows. Given a training data set $(x_i, c_i)_{i=1}^N$, the goal is to find a generative rule for new trips. Here $x_i = (s_{i1},s_{i2},\ldots,s_{in_i})$ is a speed trajectory associated with a vehicle trip $i$ and $c_i$ is the context of the trip and encodes the characteristics of the route taken and vehicle attributes used for the trip. Typically, GPS data  is used for training data set.

Current trip generation approaches \cite{1trip, powertrains,Yang21,Vatanparvar19} rely on Markov Chain model to generate a stochastic speed profile. Typically, the speed measurements $s_t$ are binned (discretized) and then transition probability matrix (look-up table)  $P_{ij} = p(s_{t+1} = j\mid s_t=i)$ is calculated by counting relative frequencies inside each of the bins. Here $P_{ij}$ is the probability that speed value at time $t+1$ is equal to $j$ conditioning on speed value is equal to $i$ at time $t$. Although this approach leads to realistic speed profiles, it does not allow to include other conditional variables into the generation process. However, for accurate energy analysis, we need to condition the speed generation on  route attributes (traffic conditions, intersection types) and vehicle characteristics (mass, torque, power-train). For example, if the vehicle reaches a stop sign at time $t+1$ then we need to have $p(s_{t+1} = 0 \mid s_t, c_t = \text{vehicle reached stop sign}) = 1$. Thus,  a more practical problem is to generate speed samples conditional on context variables that encode those characteristics of the trip and vehicle. The problem then is to estimate the transition kernel 
\begin{equation}
\label{eq:problem}
    p(s_{t+1}\mid s_t, c_t).
\end{equation}
  
Each conditional variable (component of vector $c_t$) adds a dimensionality to the transition matrix and then frequency based estimates become unreliable. There is simply not enough data in each of the bins to accurately estimate the frequency. Some bins in higher dimensions will be empty.  

In this article, we explore the problem of generating speed trajectories $s_1,s_2,\ldots,s_t$ using neural network models, that provide a more viable approach to using look-up tables in higher dimensional generation problems.  Our work is motivated by the success of deep learning (DL) models in sequential data analysis (prediction and generation), such as speech recognition \cite{speech}, music \cite{crnngan,music2}, audio \cite{melnet} and text \cite{stega} generation, as well as medical time series  analysis \cite{rcgan}.

We model this conditional distribution using neural network models, that can be viewed as an alternative to traditional look-up tables used for speed generation. Our approach relies on  Generative Adversarial Networks (GANs) \cite{gan} and normalizing flows deep leaning generative models. Although those models were designed to generate a sample in one step, the temporal nature of our data and different length of samples require sequential generation. We make the use of those generative models possible by mapping each speed sequence $x$ into a latent space. Although trips are of different lengths, their latent representation always has the same number of elements. The latent space representation is calculated by sequentially filtering the speed trajectory using a recurrent neural network. 

Our non-linear latent feature generative model takes the form 
\begin{align}
  x = & \xi(\phi)\label{eq:model1}\\
  \phi \mid z,c, \theta_g & = G(z,c, \theta_g) \label{eq:model2}\\
  z & \sim p(z).\label{eq:model3}
\end{align} 
Here $\phi$ is a latent space representation of a trajectory $x$. Tran forming original speed trajectory $x$ to a latent space representation $\phi$ allows for the relations between the context variable $c$  and new samples $x$ to be modeled by a probabilistic model $G$.  The context variable $c$ encodes the attributes of the vehicle type (engine, pertain, mass), as well as the route attributes (road types, congestion patterns)

A new sample $x$ is generated using an implicit generative model that is defined as the deterministic map $G(z,\phi, \theta_g)$, which transforms a sample of a random variable $z$, that follows some known distribution $p(z)$, such as Normal, to  $\phi$ that follows the target distribution to be modeled. The decoder map $\xi$ then transforms $\phi$ back to the space of speed trajectories. The generator map $G$ is parametrised by vector $\theta_g$. Unlike traditional statistical models, our approach does not assume a specific parametric distributional model that can be estimated using maximum likelihood. We do not make an implicit assumption about the density function. Our model only indirectly interacts with the target density and allows to draw new  samples. 

The map $\xi$ is modeled as a recurrent neural network that allows to capture temporal dependencies in the speed trajectories \cite{speech,crnngan,stega,rcgan}. The generative function $G$ is modeled using generative adversarial network or  a normalizing flow neural network. 

In this paper, we use this general latent-feature framework and extend several traditional generative models. Further, we compare those extensions on the task of speed trajectory generation. We demonstrate our approach on generating passenger vehicle trips in Chicago area. 

\section{background and related work}\label{sec:related}
Although the problem of trip generation is not well studied, there is a large amount of literature on the related problem of trip prediction and classification \cite{warren2019clusters}. There are similarities between those two and our work builds on some models proposed in the trip prediction literature.  The problem of trip prediction is that given the observed speeds from start to the present moment, the next state of the vehicle in the future, e.g., speed, acceleration, or direction of the movement, is predicted. The output of the generating model, on the other hand, is the whole trip with specific characteristics. 

In \cite{short-term}, the authors try to predict the vehicle destination based on vehicle trajectories. They use a hierarchical Markov model that uses the trip information to predict destination. Then they use these results as an input to a machine learning process along with trip information, GIS Land Use Data, participant demographic information and trip purpose estimation for destination. However, this work develops a predictive machine learning model, not a generative one. The authors of \cite{social} attempt to predict the next motion of  a vehicle in a multi-agent environment, by using the power of recurrent neural networks (RNN) in autoencoder-based approach. They use a RNN-based encoder to map the original input to a reduced dimension latent space, then feed the latent space to the LSTM based encoder. The model predicts the multi-modal distributions of the vehicle's future motions. The LSTM in the model is used to capture the temporal features of the input. In \cite{bayesian}, a Bayesian neural network is proposed for long-term trajectory predictions in order to avoid collisions in the coming seconds. Several other RNN-based approaches were recently propose for predicting coordinates of the other vehicles  \cite{grid} and trajectory prediction \cite{sequence, ann-fnn}. More traditional statistical approaches to analyze speed data  measured at a fixed location rely on particle filtering \cite{johannes2009particle,polson2015bayesian,polson2017bayesian}, traditoinal time series models \cite{williams2001multivariate,tan_aggregation_2009} and feed forward neural networks \cite{polson2017deep,dixon2019deep}.

Previously developed generative models relied on strong assumptions of the nature of the speed profile, such as the data being periodic \cite{ann-fnn}. However, energy estimation algorithms are sensitive to inaccuracies in the input speed data and require speed profiles mimic to be realistic and of high temporal resolution, of at least 1Hz. The previously proposed non-parametric techniques \cite{1trip, powertrains,Yang21,Vatanparvar19} do not make those restrictive assumptions, but are not scalable to hire dimensional settings. 

Our work builds on generative models developed for text, speech, and music, as well as other vehicle trajectory prediction problems. In recent years, many researchers worked on time-series generation and prediction by developing deep learning models. In this section, we first explore the most dominant deep learning models for sequence generation, and then present our extension developed for vehicle trajectory prediction and generation. 

The rest of the paper is organized as follows: Section \ref{sec:related} covers the background and explores the state-of-the-art  techniques. The data set used in this paper is described in \ref{sec:data}. Section \ref{sec:method} explains the generative model that we use for the vehicle trajectory generation task. Section \ref{sec:results} evaluates the results of the generative model. Finally, Section \ref{sec:conclusion} concludes with discussion and future research directions.

\subsection{Time-Series Generation with Deep Learning Models}
In this section, we provide specifics of the models that are used in our non-linear latent feature generative model given by Equations (\ref{eq:model1}) - (\ref{eq:model3}). First, we introduce recurrent neural networks that are used to map time series observation $x$ to latent feature $\phi$ (encoding) and its inverse for mapping from latent space back to speed trajectory space (decoding). Then we introduce autoencoder that combines encoder and decoder into one model and allows to jointly train both of the networks. Further, we introduce  Generative Adversarial Networks (GAN),  and Normalizing Flow models we used to define the map $G$. GANs and normalizing flows were originally designed to generate a sample in one step, the variable length of a speed trajectory and the nature of conditional variables $c$ prevents us from using the original architectures. We extend those generative models by combining them with latent feature maps.

\subsubsection{Recurrent Neural Networks}
Both GANs and Normalizing flows models can implicitly estimate the probability density function. However, a typical feed forward architectures used for those networks are not capable of learning the temporal patterns. We propose using the recurrent neural networks. The main difference between the feed-forward  and recurrent architectures is that an RNN has a memory variable that is used to model temporal relations present in data. 
We use recurrent neural networks to define encoder $\eta(x) = \phi$ and decoder $\xi(\eta) = x$  maps. Recurrent neural network $\eta(x)$ sequentially analyzes the speed observations. Each iteration of a recurrent neural network outputs a nuance memory variable $h_t$, called a hidden state, that is used as an input for the next iteration $t+1$. Therefore, we can formulate an RNN as follows \cite{unreasonable}:
\begin{equation}
h_t=f_1(\theta_{hh}h_{t-1}+\theta_{sh}s_t)
\end{equation}

\begin{figure}[t]
  \centering
    \includegraphics[width=0.7\columnwidth]{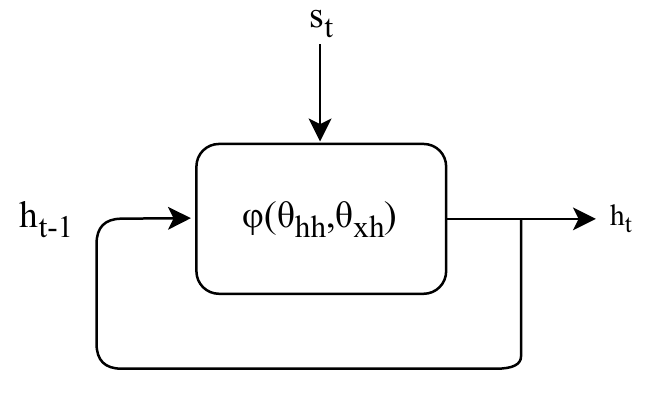}
  \caption{Block diagram of a Recurrent Neural Network}
  \label{fig:rnn_block}
\end{figure}

Where $h_t$ is the output of the RNN at time $t$, $s_t$ is the input at time $t$,  $\theta_{hh}$ and $\theta_{sh}$ are the weights of the network corresponding to the output and the input, respectively. A block diagram of a RNN is shown in Figure \ref{fig:rnn_block}. Then the latent representation of input $x = (s_1,\ldots,s_n)$ is simply the hidden variable calculated at the last step of RNN $\eta(x) = h_n = \phi$.

The decoder, which is the inverse of encoder, sequentially generates the speed trajectory as follows:
\begin{equation}
(s_t,h_t) = f_2(\theta_{h}h_{t-1}+\theta_{s}s_{t-1}),
\end{equation}
we initialize the decoder by setting $h_0 = \phi$ and $s_0 = 0$.

RNNs are powerful models in sequence generation tasks, such as text \cite{ts,stega,order} and speech recognition and synthesis \cite{speech,investigating}. RNN were shown to be efficient for trajectory prediction and generation \cite{dynamic,modeling,Australia}. Thus, we consider an RNN-based model for trajectory generation problem. 

One of the practical issues when using RNNs is the problem of vanishing gradient, meaning that at some stage of the training process, rounding errors dominate the calculations.

\subsubsection{LSTMs}
Long-short term memory (LSTM) models address the problem of vanishing gradients by introducing additional filters, called gates, at each recurrent cell that filter out unrelated parts of the hidden state \cite{lstm}. An LSTM cell is defined as

\begin{align*}
  f_t&=\sigma(W_{sf}s_t+W_{hf}h_{t-1}+b_f)\\
  i_t&=\sigma(W_{si}s_t+W_{hi}h_{t-1}+b_i)\\
  o_t&=\sigma(W_{so}s_t+W_{ho}h_{t-1}+b_o)\\
  g_t &= \tanh(W_{gs}s_t+W_{gh}h_{t-1}+b_c)\\
  u_t&=f_t\odot u_{t-1} + i_t\odot g_t \\
  h_t&=o_t\odot \tanh(u_t).
\end{align*}

\begin{figure}[t]
  \centering
    \includegraphics[width=\columnwidth]{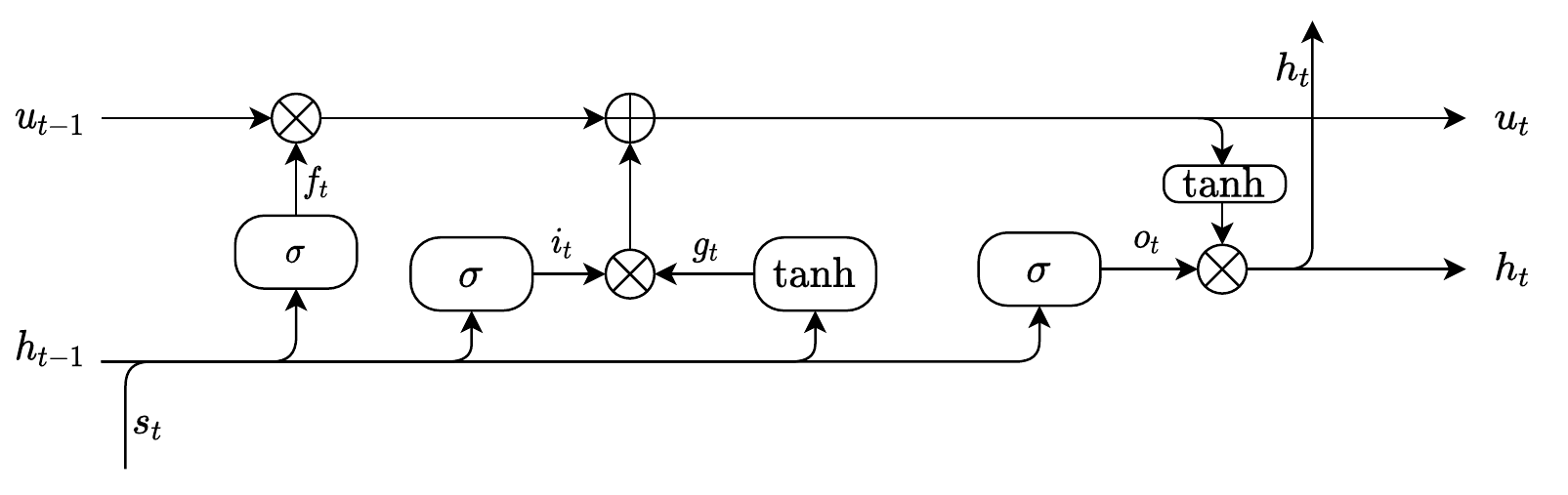}
\caption{ Block diagram of a LSTM}
  \label{fig:lstm}
\end{figure}

Here $x\odot y$ : element-wise product and $\sigma(x)$ is the sigmoid function.
Figure \ref{fig:lstm} shows the block diagram of an LSTM cell \cite{lstm-block}. Gating vectors provide the information required for the cell memory to  update,forget, and output its state by  $f_t$, $i_t$ and $o_t$ respectfuly. The forget vector $f_t$, determines if the cell state should reset or restored.  Equations for $u_t$ and $h_t$ are then responsible for updating the cell state and output. 

\subsubsection{Autoencoder}
Autoencoders (AEs) are deep neural networks that map the input to itself via a bottleneck structure, which means the model $\phi =  \eta(x)$  aims to concentrate information required to recreate input $x$ using a latent vector $\phi$. We can think of this latent vector as a low-dimensional representation of $x$. Autoencoder has two deterministic maps, the reduction map $\phi =  \eta(x)$  and reconstruction (decoder) map $\hat x = \zeta(\phi)$. The parameters of both of the maps are estimated jointly by minimizing the least squares loss
\begin{equation}
\label{eq:AE-loss}
\sum_{i=1}^{N}(x_i - \zeta(\eta(x_i)))^2 \rightarrow \mathrm{minimize}. 
\end{equation}

\subsubsection{Generative Adversarial Networks}
Generative Adversarial Networks (GANs) allow to learn the implicit probabilty distribution over $x$ by defining a deterministic  map $x = G(z,\phi, \theta_g)$, called  generator. The basic idea of GAN is to introduce a nuisance neural network $D(\phi, {\theta_d})$, called discriminator and parametrised by $\theta_d$ and then jointly estimate  the parameters $\theta_g$ of the generator function $G(z,\phi, \theta_g)$ and the . The discriminator network is a binary classifier which is trained to discriminate generated and real samples $x$ and the parameters are found by minimizing standard cross-entropy loss, traditionally used to estimate parameters of binary classifiers
\begin{equation}
\label{eq:gan_loss}
  \begin{split}
  \MoveEqLeft   
    J(\theta_d,\theta_g) = -\dfrac{1}{2}E_{\phi}[\log D(\phi, {\theta_d})]\\
   &-\dfrac{1}{2} E_{z}[\log(1- D(G(z,\phi, \theta_g), {\theta_d}))].
  \end{split}
  \end{equation}
To calculate the first term, the expectation with respect to $x$, we just use empirical expectation calculated using observed training samples. Next, we need to specify the cost function for the generator function. Assuming a zero-sum scenario in which the sum of the cost for generator and discriminator is zero, we use the mini-max 
estimator, which jointly estimates the parameters $\theta_d$ (and $\theta_g$ as a by-product) and is defined as follows:
\begin{equation}
\begin{split}
\MoveEqLeft   
 \min_{\theta_g}\max_{\theta_d} J(\theta_d,\theta_g)
\end{split}
\end{equation}
The term adversarial, which is  misleading, was used due to the analogy with game theory. In GANs the generator networks tries to ``trick'' the discriminator network by generating samples that cannot be distinguished from real samples available from the training data set. 

\begin{figure}[t]
  \centering
    \includegraphics[width=\columnwidth]{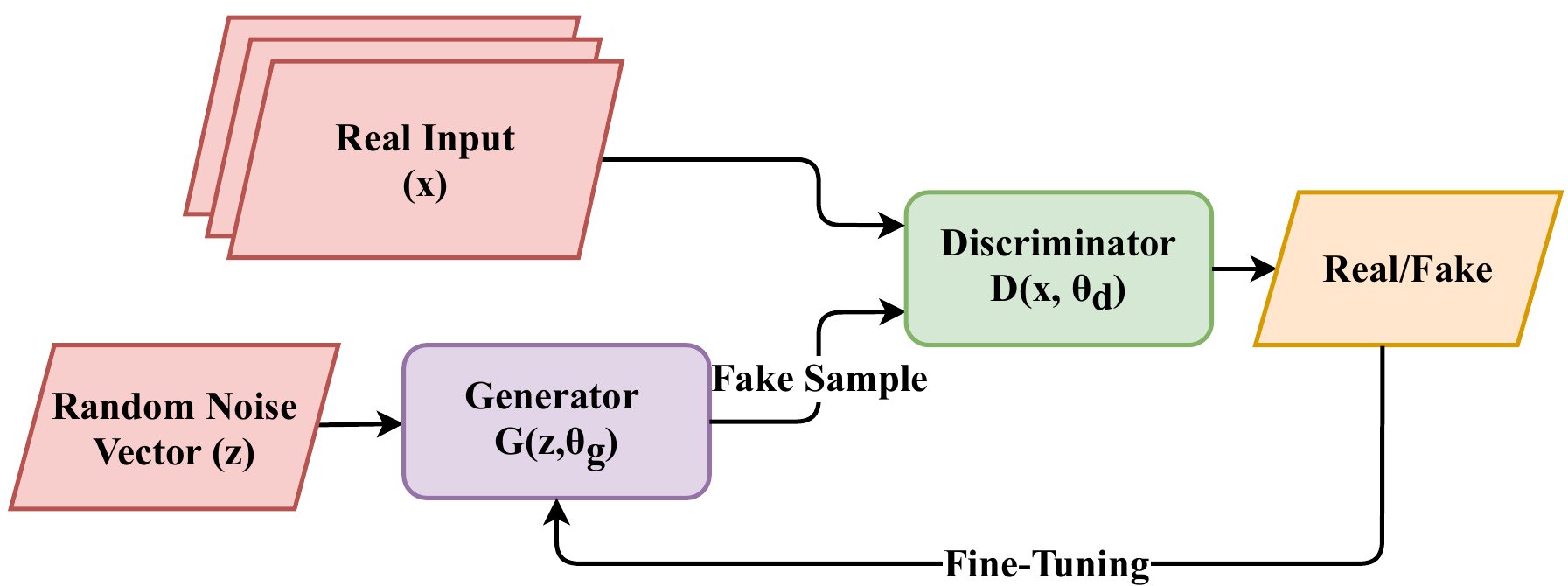}
  \caption{Block diagram of a Generative Adversarial Network}
  \label{fig:gan_block}
\end{figure}

Figure \ref{fig:gan_block} shows the block diagram of a typical GAN network. GAN architectures with feed-forward neural networks as generator and discriminator have proven to be powerful models for image  generation \cite{progressive,visually,can}. 

In our architecture, both the generative and discriminative networks are modeled by a feed forward neural network. A feed forward neural network is a composite map. The generation map is defined by
\begin{equation}
\label{eq:reduction_map}
G(\phi)  = \left ( f_{w_n,b_n} \circ \ldots \circ f_{w_1,b_1} \right ) (\phi),   
\end{equation}
where each layer of the architecture $f_{w_l,b_l}$, $l=1,\ldots,n$ is a semi-affine activation rule defined by
\begin{equation}\label{eq:deep-function}
f_{w_l,b_l} (\phi) =  f \left ( \sum_{j=1}^{N_l} w_{lj} \phi_j + b_l \right ) = f ( w_l^T \phi_l + b_l ).
\end{equation}
Here, $N_l$ denotes the number of activation units at layer $l$. The weights $w_l \in \mathbb{R}^{N_l \times N_{l-1}}$ and offset $b\in \mathbb{R}$ need to be learned from training data.

\subsubsection{Normalizing Flows}
Normalizing flows provide an alternative approach of defining a deterministic map $x \mid \phi, \theta_g  = G(z,\phi, \theta_g)$ that transforms a univariate random variable $z\sim p(z)$ to a sample from the target distribution $G(z,\phi, \theta_g)= x \sim F(x)$. If transformation $G$ is invertible ($G^{-1}$ exists) and differentiable, then the relation between target density $F$ and the latent density $p(z)$ is given by \cite{NF}:

\begin{equation}
    F(x) =p(z)\left|\det\frac{\partial G^{-1}}{\partial z}\right| = p(z)\left|\det\frac{\partial G}{\partial z}\right|^{-1}
\end{equation}
where $z = G^{-1}(\phi)$. A typical procedure for estimating the parameters of the map $G$ relies on maximizing the log-likelihood
\begin{equation}
\label{eq:max_likelihood}
  \log p(z) + \log \left|\det\frac{\partial G^{-1}}{\partial z}\right|
\end{equation}

The normalizing flow model requires constructing map $G$ that have tractable inverse and Jacobian determinant. It is achieved by representing $G$ as a composite map
\begin{equation}
\label{eq:NF1}
G = T_k \circ \cdots \circ T_1,
\end{equation}
and to use simple building block transformations $T_i$ that have tractable inverse and Jacobian determinant.

The likelihood for such a composite map is easily computable. If we put $z_0 = z$ and $z_K = x$, the forward evaluation is then
\begin{equation}
\label{eq:NF2}
z_k = T_kz_{k-1}, \text{ for } k=1,\ldots,K,   
\end{equation}

and the inverse evaluation is 
\begin{equation}
\label{eq:NF3}
  z_{k-1} = T^{-1}_k(z_k), \text{ for } k=1,\ldots,K.    
\end{equation}

Further, the Jacobian is calculated as the product of Jacobians
\begin{equation}
\label{eq:jacobian}
\left|\det\frac{\partial G^{-1}}{\partial z}\right| = \prod_{k=1}^{K}\left|\det\frac{\partial T_k}{\partial z_{k-1}}\right|^{-1}.    
\end{equation}

\subsubsection{Hybrid Models}
Combining a recurrent neural networks with other architectures has been a fruitful approach when modeling temporal data. Combination of an RNN and convolutional feed-forward networks was effective when modeling periodic time-series is developed in \cite{lstnet}. \cite{rcgan} proposes a conditional RNN-based generative adversarial network to generate medical time-series data. A LSTM-based GAN is proposed in \cite{crnngan} to generate music data that sounds good.  \cite{sequence,traphic,social,generalizable,imitating} use hybrid models to predict the trajectories of the vehicle or the vehicles in surrounding environments. To use the power of generative neural nets such as GANs, Autoencoders, and Normalizing flows for time series data, we develop a hybrid architecture that combines those with recurrent neural networks. We build on similar approaches that were developed  to perform  predictive or generative tasks for temporal data \cite{social,rcgan,crnngan,traphic,hybrid,time-gan,sequence,lstnet,pixelsnail,multi}. 


\begin{figure*}[!htb]
  \centering
    \includegraphics[width=\textwidth]{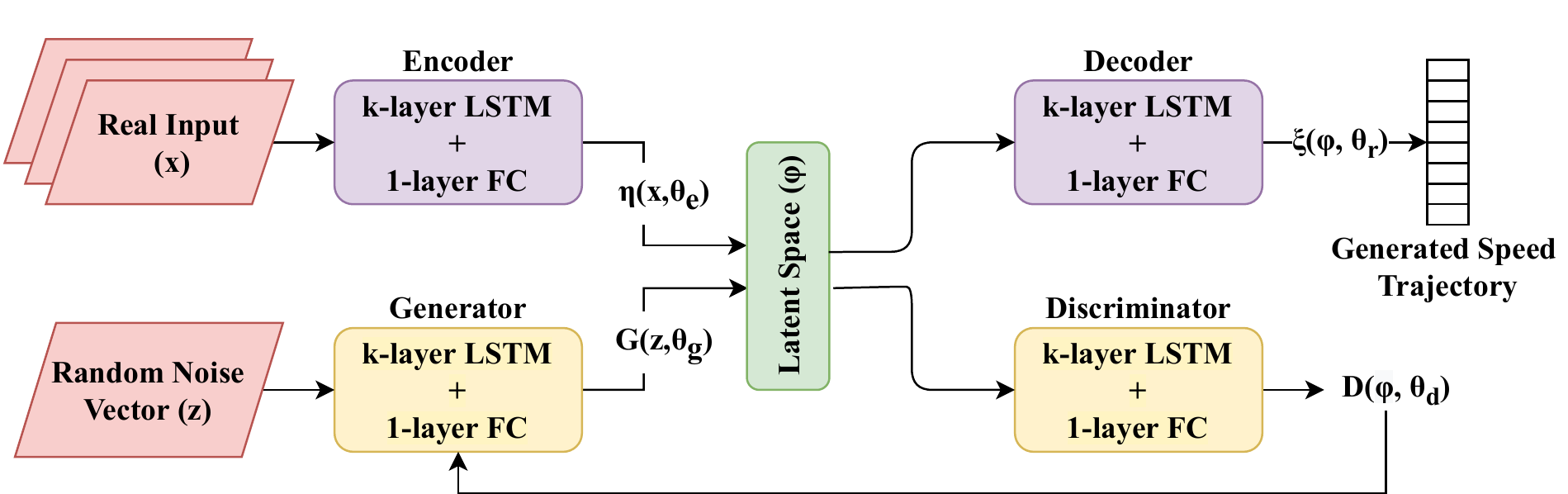}
  \caption{Block diagram of the utilized model, an AE/GAN hybrid network, constructed of LSTM and fully connected (FC) layers}
  \label{fig:time-gan}
\end{figure*}

\section{Chicago Data}\label{sec:data}
\begin{figure}[H]
  \centering
    \includegraphics[width=0.85\columnwidth]{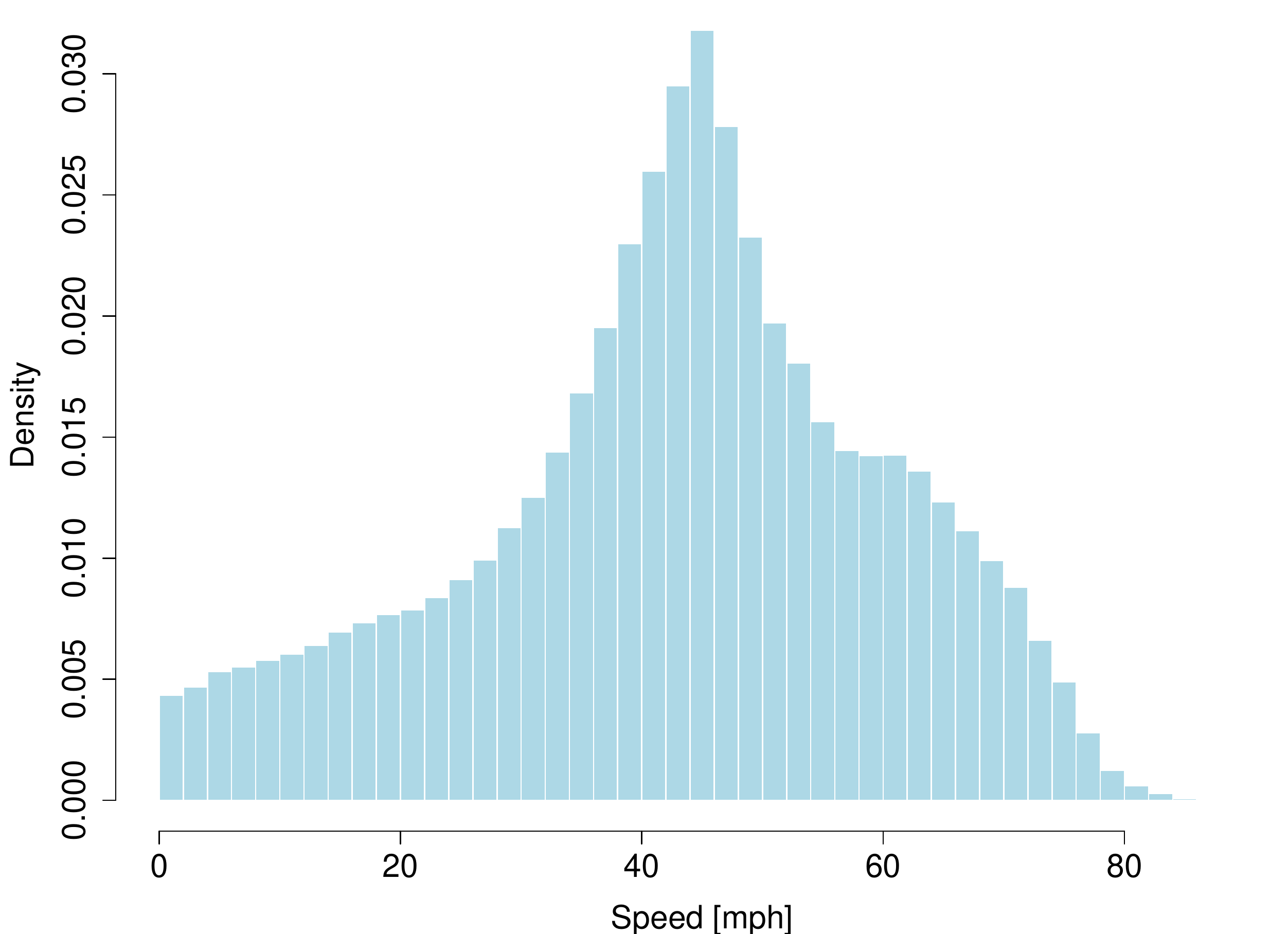}
  \caption{Distribution of the speeds}
  \label{fig:data}
\end{figure}

We use speed trajectory data collected from GPS sensors at 1 second interval for training  and validating our models. The data was collected as part of traveler's survey conducted by by the Chicago Metropolitan Agency for Planning in the Chicago metropolitan area and was provided to us by Argonne National Lab (ANL). Each data point  contains the timestamp, the speed value, and the longitude and latitude coordinates where the speed was measured. The speed values in the data set are in the range [0-35] m/s. There are 1.9 million data points in the training sets, with the trip lengths varying between 100 and 6330 seconds. In this paper, we concentrate on the temporal sequence of captured speed values. The dataset includes multiple trips with different lengths and starting and ending point.  Each trip $x_i$ is thus a sequence of speed measurements $x_i = (s_{i1},\ldots,s_{in_i})$. Figure \ref{fig:data} shows the density distribution of the speeds.
As we can see in the density plot, the peak density of the speeds in the middle range at about 40 mph, which corresponds to driving on arterial roads or mildly congested highways.

\begin{figure}[H]
\centering
\includegraphics[width=\linewidth]{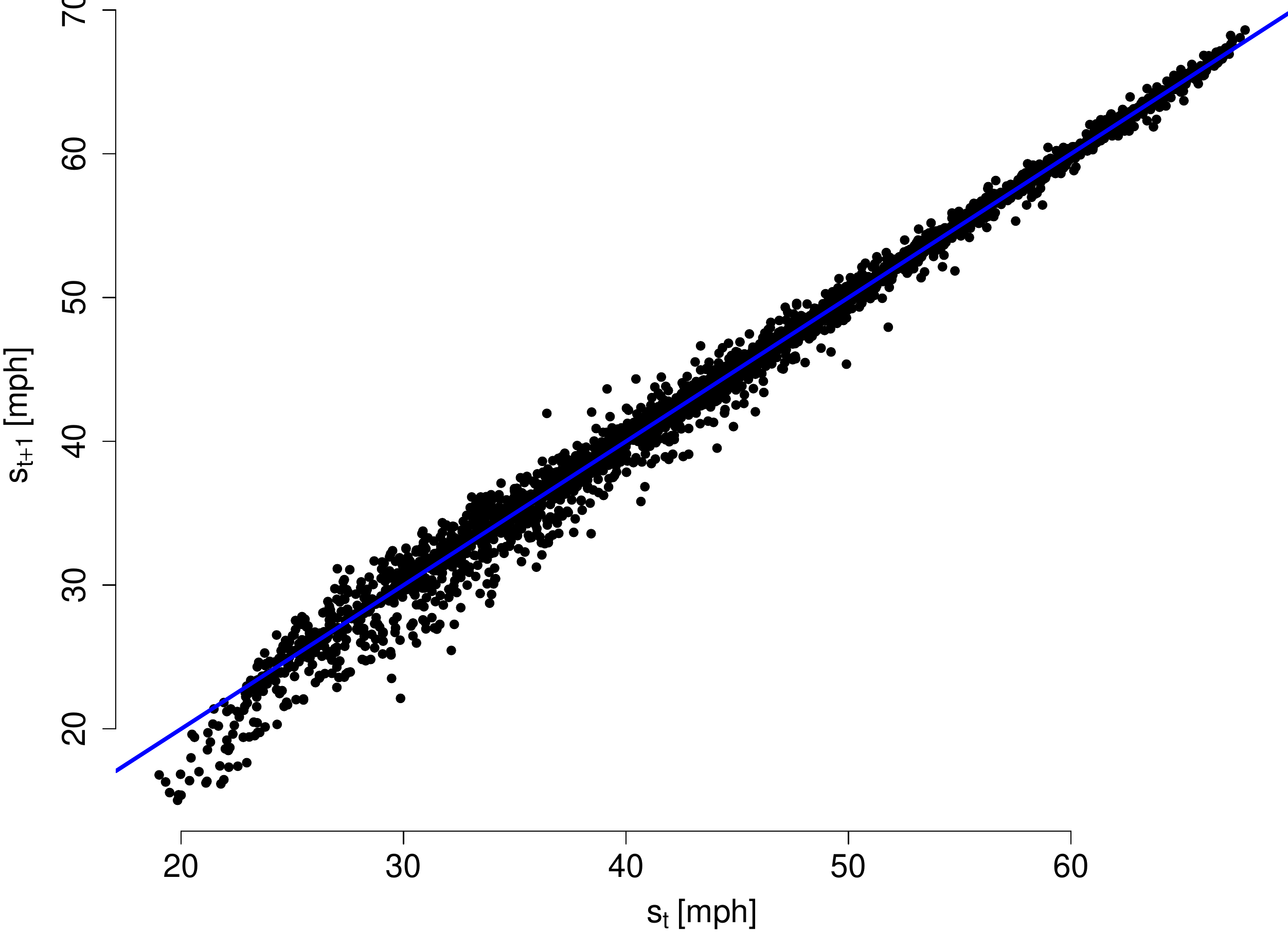}
\caption{Scatter plot of two consecutive speed observations}
\label{fig:speed-scatter}
\end{figure}

Our models generate data sequentially and use previous speed observation $s_t$ as one of the inputs to generate the next speed $s_{t+1}$. Figure \ref{fig:speed-scatter}  shows the scatter plot along with a linear regression line.
The pattern shown in Figure \ref{fig:speed-scatter} motivates the use of non-linear conditional models for sequential speed generation. Our attempts to build a linear model by using heteroskedastic error models and data binning did not lead to successful outcomes.

\section{Methodology}\label{sec:method}
In this section, we describe the extensions to GANs that we used to build our generative model given by Equations (\ref{eq:model1}) - (\ref{eq:model3}). We start my describing an unconditional model that does not depend on the context variable $c$. The unconditional model generates speed profiles that have the same density as the dataset. Further, we describe the conditional model that allows to account for the road characteristics. 

\subsection{Unconditional Model}
Our work builds on a GAN model \cite{time-gan} and we use recurrent neural network architectures for both generator and discriminator so that our model captures the temporal patterns in the speed data.

We train our model using the set of GPS trips. Joinlty training a GAN with auto-encoder RNN architectures is challenging. When training a GAN, we are training two competing neural networks simultaneously, and optimizing one may be at the expense of the other neural network. Finding a good architecture is challenging. Although there are automated techniques to adjust neural network configurations \cite{stanley2002evolving,baker2016designing}, we did not find those approaches useful and used a random search to identify the specifics of the architectures. The block diagram of the model developed in this paper is shown in Figure \ref{fig:time-gan}.

As we mentioned above, our model is a combined autoencoder/GAN. The autoencoder consists of RNNs, for the same reason we used RNNs for GANs. Therefore, the encoder is enabled to capture the temporal features in the latent space as well. The combined AE/GAN is trained simultaneously, performing the tasks of encoding and generating. 

The input of the GAN part of the model is then the latent space, it will act as the real input to discriminator, and the generator input, the noise vector $z$ is in the same dimension of the latent space, and the generated output is also in the latent space dimension which is also an input to the discriminator. When generating real trajectories, the generate output that is in latent space dimension is reconstructed by the decoder to produce a sample in the original input dimension. In other words, in Equation (\ref{eq:gan_loss}), $\phi$ is the latent space produced either by encoding the real sample $x$, or the generated sample in the latent space domain.

\subsection{Conditional Model}
\begin{figure}[b]
  \centering
    \includegraphics[width=0.85\columnwidth]{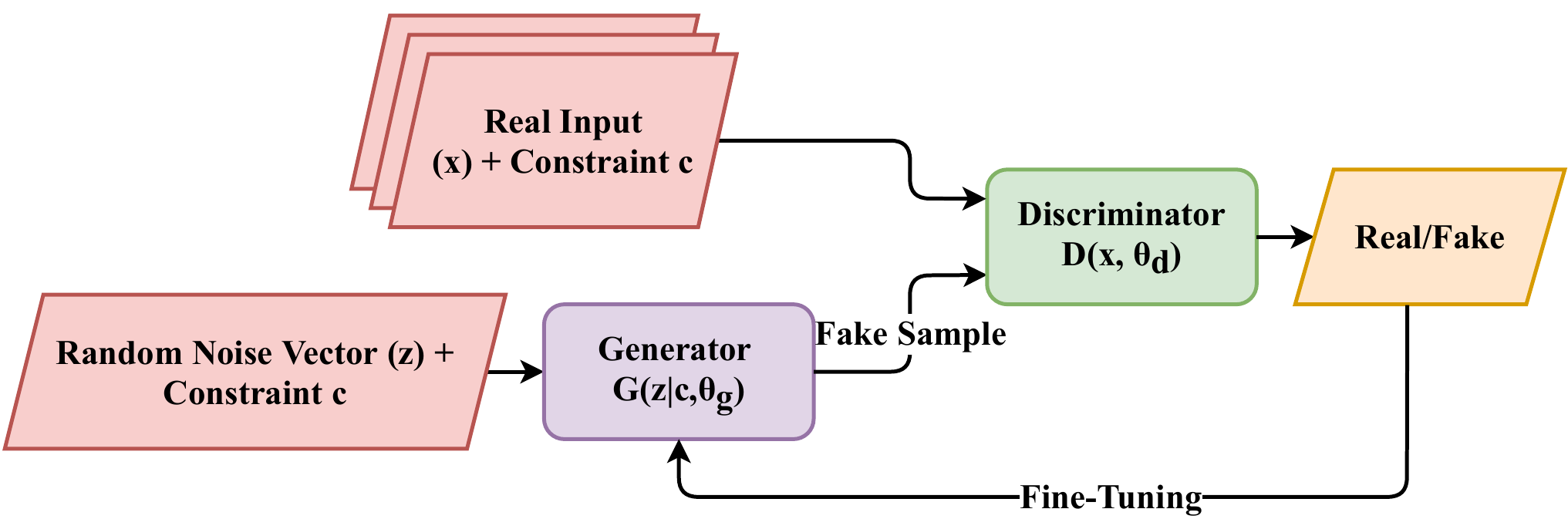}
  \caption{Conditional GAN block diagram}
  \label{fig:con-gan-block}
\end{figure}

One of the limitation of the previously used Markov-chain approaches is the inability to incorporate constraints into generation algorithms. Constraints come from the attribute of road segments vehicle is traveling on. For example, the vehicle has to stop at a stop-sign controlled intersection.  Flexibility of neural network architectures allows us to account for those constraints. We extend a conditional GAN that was first introduced by \cite{con-gan} and we add the class label to both the discriminator and generator, by concatenating them to the real input and the noise vector, respectively. In this way, the model can learn to condition its generated sample based on the provided class. Figure \ref{fig:con-gan-block} shows the modification to the plain GAN that enables the model to generate samples belonging to a desired class. The objective function of the GAN would change as follows:
\begin{equation}
\begin{split}
  \MoveEqLeft   
  J(\theta_d,\theta_g) = E_{\phi}(\phi)[\log D(\phi|c)]\\
  &+ E_{(z)}[log(1-D(G(z|c)))],
\end{split}
\end{equation}
where $c$ is the desired class, or the condition we want the GAN to learn. Therefor the loss functions of both discriminator and generator are conditional on $c$.

For a recurrent neural network, we can apply the same approach. We can condition the generation on a constraint that is not dependent on time  to both the discriminator and generator input by adding the conditional variable $c$ at each step to the input and the noise.

\section{Experimental Results}\label{sec:results}
In this section, we explore the sequences generated by the proposed architectures. We apply our model to generating speed trajectories for passenger vehicles and use data from Chicago metropolitan area for training. We compare empirical histograms of generated and measured samples to validate our model. The best architecture for these networks, i.e., the number of layers, hidden nodes, etc. are determined via random search experiments. We train the models described in the previous section on the speed trajectories of the Chicago area.  The number of epochs and the number of LSTM layers varies by each model. The number of nodes in each LSTM layer is 24. All the experiments have been conducted in PyTorch \cite{pytorch} library. 

We demonstrate performance of four different models. The first model uses normalizing flows architecture to model the generator $G$ from Equation (\ref{eq:model2}). The rest of the models use generative adversarial networks for the generator. 

\subsection{Normalizing Flow Generator}

\begin{figure}[H]
  \centering
  \begin{tabular}{cc}
  \includegraphics[width=0.45\linewidth]{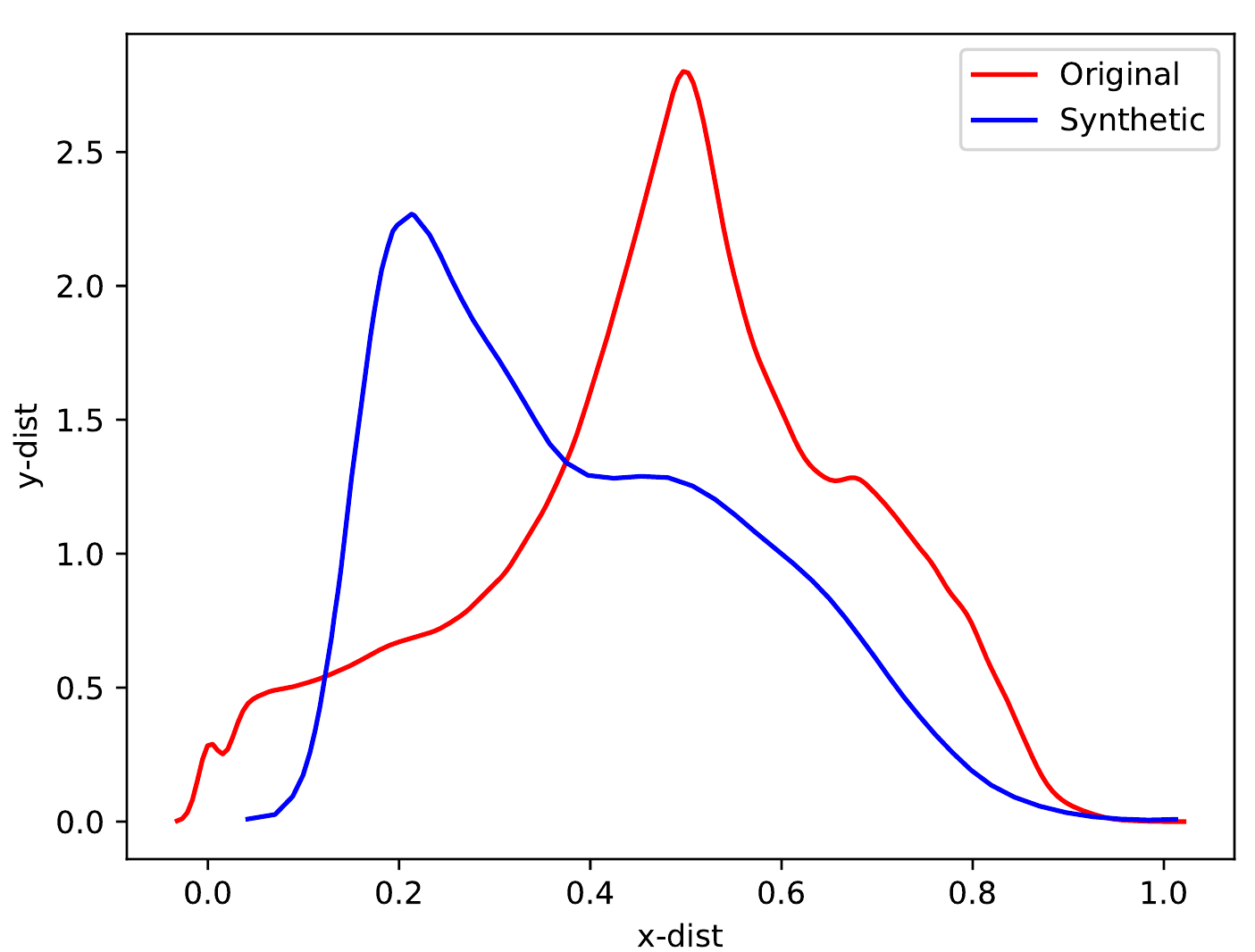} & \includegraphics[width=0.45\linewidth]{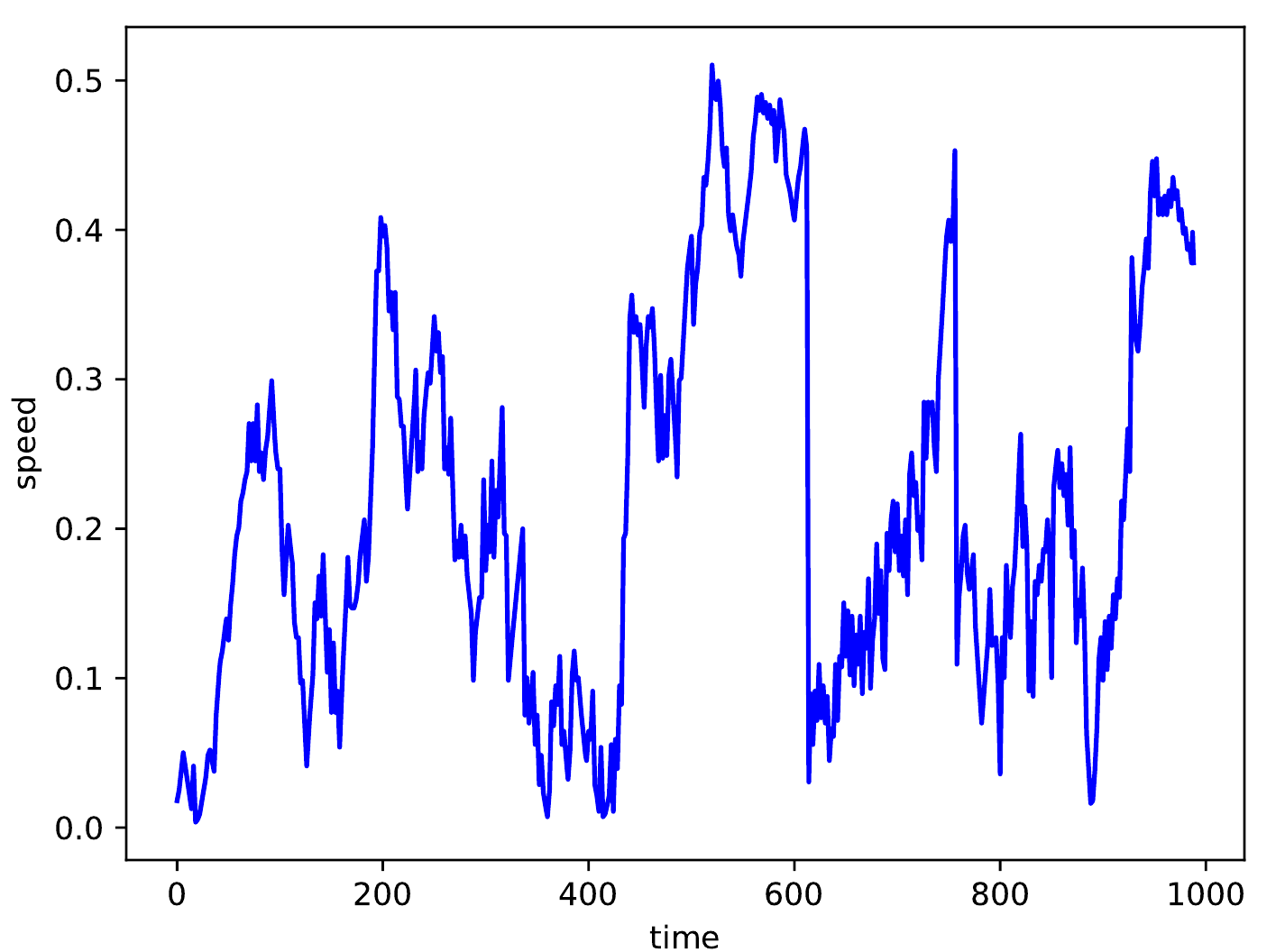}\\
  (a) Speed density plot & (b)  Sample trajectory
  \end{tabular}
  \caption{Comparison of speed  density plots for data generated with NF model (blue) and observed data (red) and sample trajectory, we see that the generated trajectories do not the same statistical characteristics of the original trajectories}
  \label{fig:NF_results}
\end{figure}
The first instance of our framework uses a Normalizing Flow (NF) generator. To use the NF model for our problem and be able to generate trajectories, we consider different intervals for the speeds. The speeds are in the range of [0,40] m/s. We divide the speeds into several bins, with 0.5 m/s intervals, so we have 80 distribution, one for every bin. To model these distributions, we train a NF for each interval, so we have 80 NF models. To use these models for generating the trajectory, we pick an interval to start, use the trained NF to sample some speeds in that interval, and randomly choose one speed from the sample distribution. Then, we check that what interval the selected speed belongs to, so we can choose the next trained NF model to get the sample distribution and generate the next speed. The generation process is described in algorithm \ref{alg::NF}. The density plots of the original and generated trajectories are shown in Figure \ref{fig:NF_results}(a). We can see that the generated trajectories have a different range and mean than the original trajectories. In Figure \ref{fig:NF_results}(b) a sample trajectory is illustrated. The synthetic trajectory has a large jump at around 400 second mark and a large drop at around 600 second mark. Those two are outlets and simply violate laws of  vehicle dynamics. Thus, the Normalizing flows does not lead to a practical model for speed generation. 

\begin{algorithm}
\caption{Normalizing Flow Generator (NFG)\label{alg::NF}}

\begin{algorithmic}[1]
\scriptsize
\State {Initiate $s = 0$}
\State{Add $s$ to Trajectory;}
\For {$i=1$ to $N$}
\State {Find the interval $j,k$ so that $s\in [s_j, s_k]$;}
\State {Use the $NFG_{jk}$ to sample $s'$;}
\State{Add $s'$ to Trajectory;}
\State {$s = s'$}
\EndFor
\State{Return Trajectory;}
\end{algorithmic}
\end{algorithm}

\subsection{One-Dimensional Unconditional RNN Generator}
\begin{figure}[H]
  \centering
  \begin{tabular}{cc}
  \includegraphics[width=0.45\linewidth]{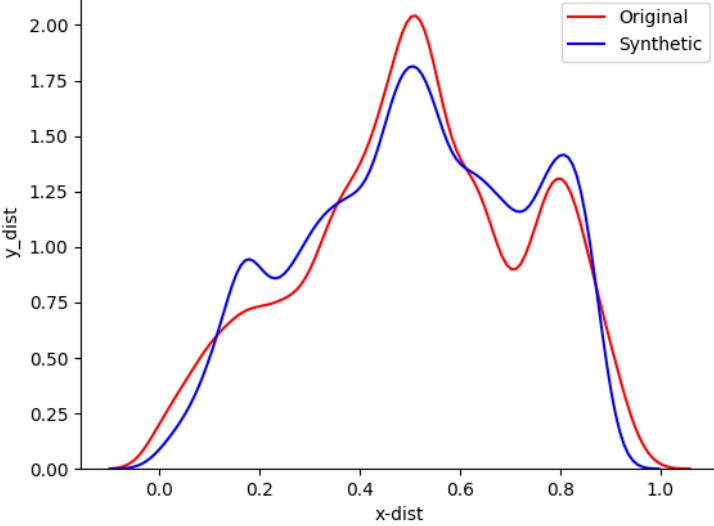} & \includegraphics[width=0.45\linewidth]{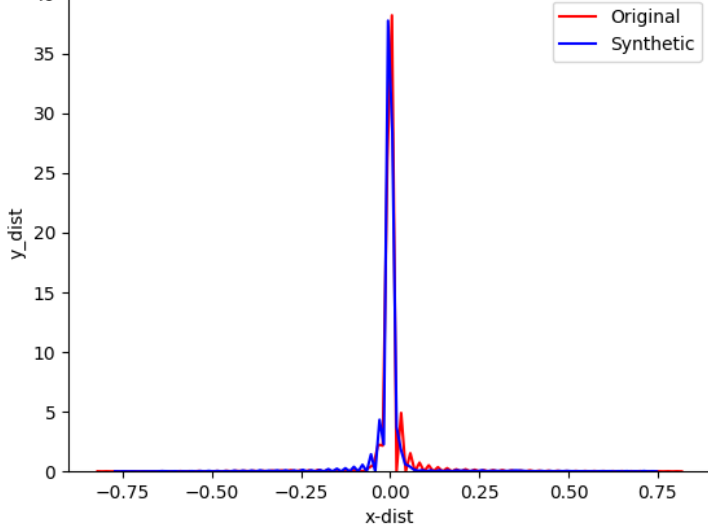}\\
  (a) Speed density plot & (b)  Acceleration density plot\\
  &
  \end{tabular}
  \caption{Comparison of speed and acceleration density plots for data generated with RNN-1D model (blue) and observed data (red). RNN-1D model was trained using early stop approach with 25000 epochs.}
  \label{fig:rnn1d}
\end{figure}

We use an RNN based model (called RNN-1D) as is an unconditional generator on uni-variate time series data.  The RNN-1D model has three LSTM layers with 24 hidden nodes for each of the Encoder, Decoder, Generator, and Discriminator networks. To train the model a batch size of 256 and Adam optimizer with the learning rate of 0.001 were used. After training the model for 25000 epochs, we generate several diverse trajectories. We plotted the probability distribution for 30000 original data points and 30000 generated data points, which are shown in Figure \ref{fig:rnn1d}(a).
As we can see in Figure \ref{fig:rnn1d}(a), the density plots of original and generated data sets are very close, as their probability distribution is similar, and the mean and range of the speed values are realistically close. Therefore, we can say that the generated trajectories are realistic. Furthermore, we  plotted the distribution of acceleration, which is illustrated in Figure \ref{fig:rnn1d}(b). The accelerations are calculated as $a_t = s_t - s_{t-1}$, and \ref{fig:rnn1d}(b) shows that the distribution of accelerations of generated and original trips also matches well.

\begin{figure}[H]
\centering
\includegraphics[width=\linewidth]{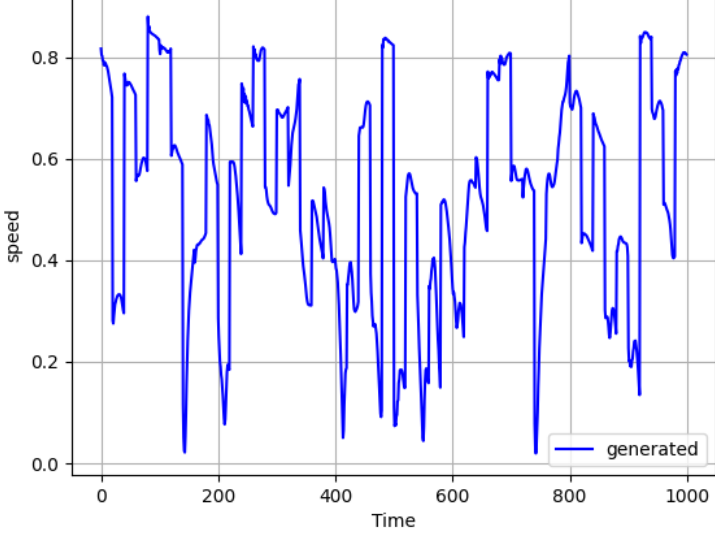}
\caption{Example of a trajectory generated with RNN-1D model.}
\label{fig:traj-timegan}
\end{figure}

Further, Figure \ref{fig:traj-timegan} illustrates a generated trajectory of length 1000.

\subsection{Three-Dimensional Conditional RNN Generator}
\begin{figure}[H]
\centering
\begin{tabular}{cp{0.45\linewidth}}
\includegraphics[width=0.45\linewidth]{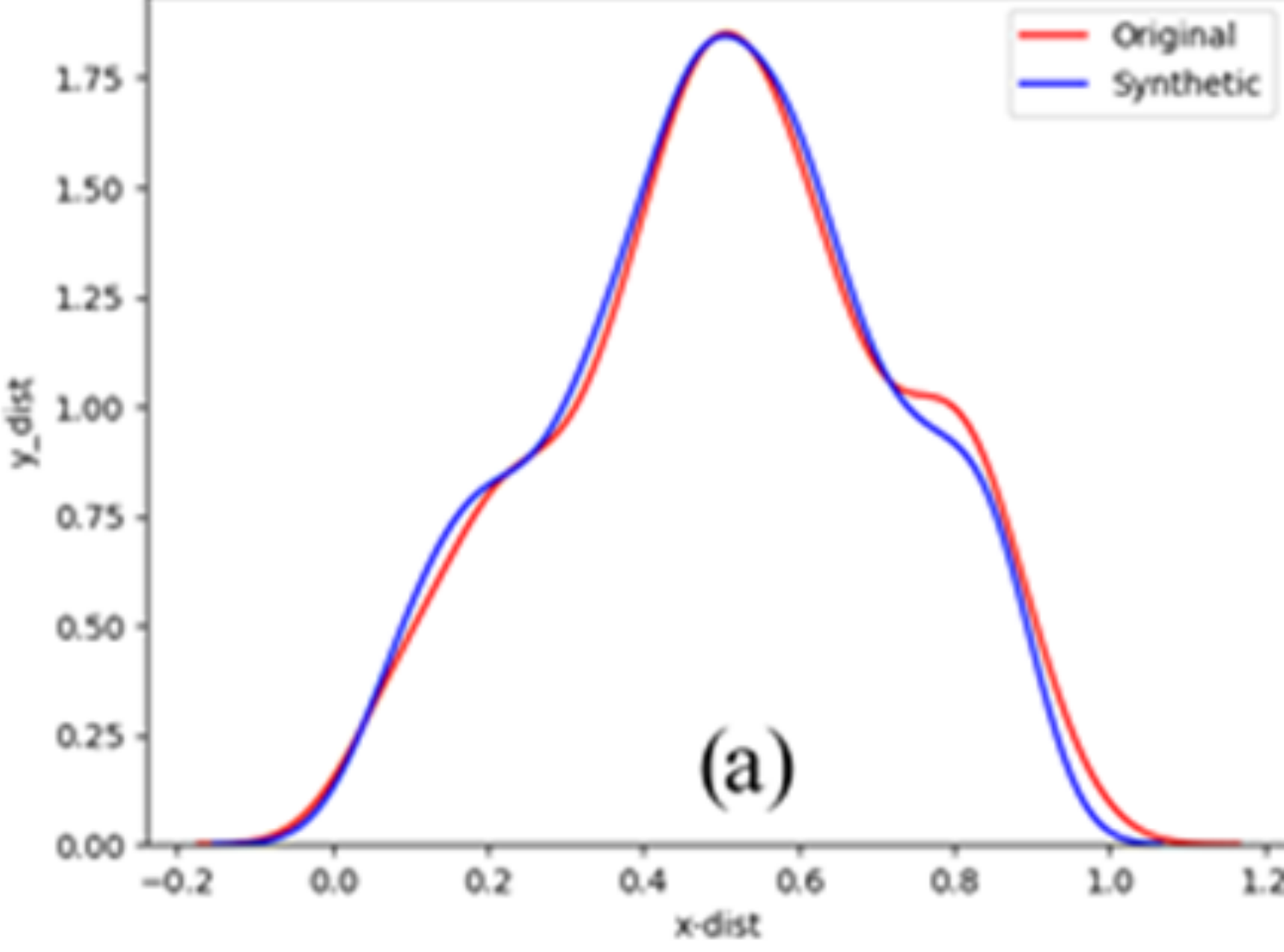} & \includegraphics[width=\linewidth]{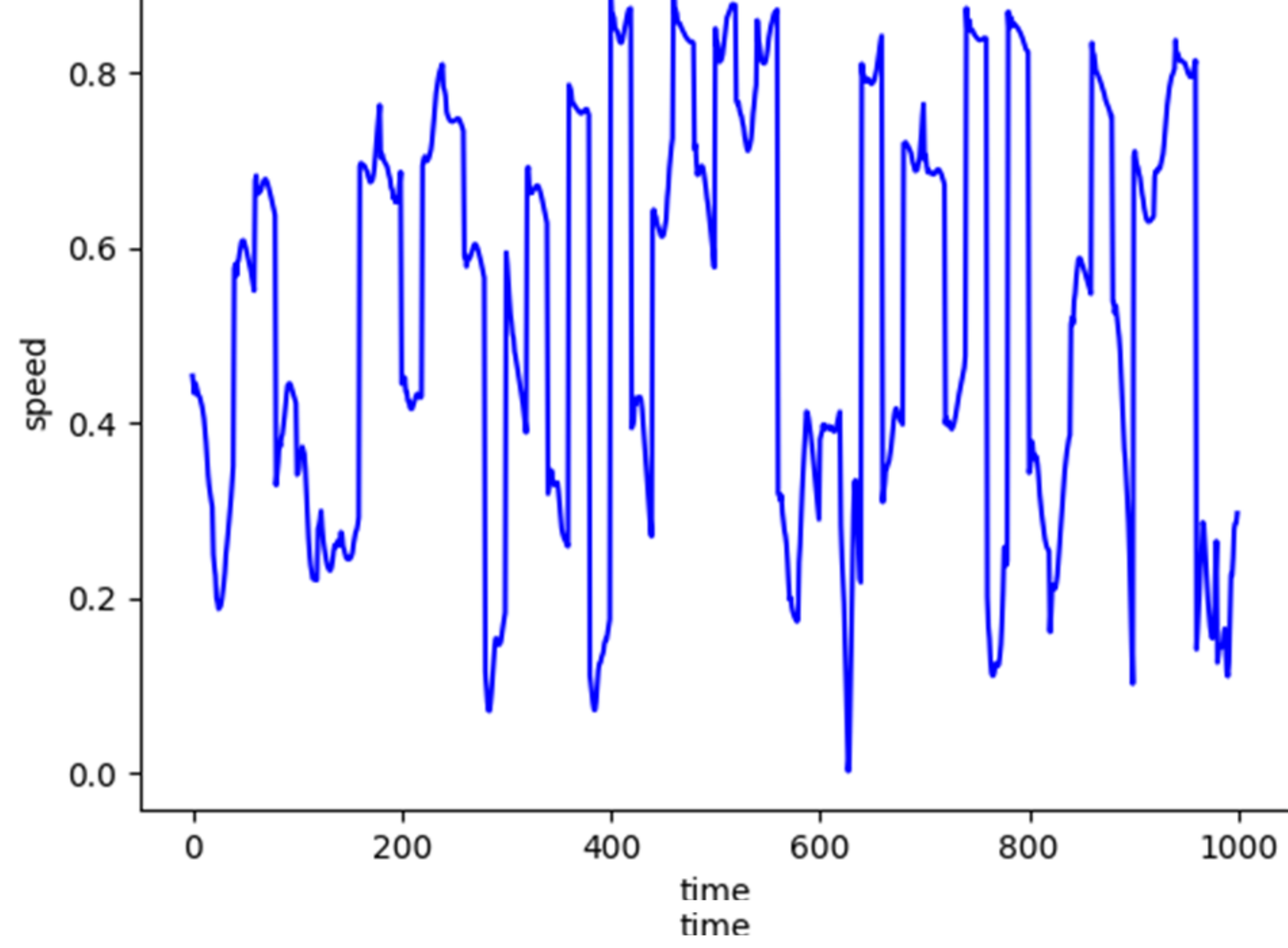}\\
(a) Speed density comparison & (b) Example of a generated speed trajectory\\
&
\end{tabular}
\caption{Speed density plot and a generated trajectory for 3D-RNN model.}
\label{fig:3drnn}
\end{figure}

We developed another model (RNN-3D) that takes the following additional inputs: the length of each trip, $L$, and the distance $d_t$ of the vehicle to zero in each moment $t$. For each trip, $x$ is the same for every entry of the trajectory, while $d_t$ can be calculated as:
\begin{equation}
\label{eq:distance_to_zero}
    d_t = L - \sum_{i=0}^{t}{s_i}.
\end{equation}
The input of the decoder model is now a multi-variate time-series $(s_t, d_t, L)$. The architecture and the learning hyper-parameters of this model is similar to RNN-1D, except we used 2 LSTM layers with 24 hidden nodes for each of Encoder, Decoder, Generator, and Discriminator networks. The density plots of original and generated speeds (30000 for each) are shown in Figure \ref{fig:3drnn}(a). As we can see, the density plots are much closer in this model, therefore, the model is benefiting from the additional input features provided. The sample trajectories generated by this model is demonstrated in Figure \ref{fig:3drnn}(b).

\subsection{Conditional RNN (C-RNN)}
Although we added the additional inputs to the RNN-3D and got better trajectories as a result of the additional information, it wasn't enough to generate trajectories to satisfy the constraints like the length of the trip. Therefore, we need to develop a model that not only can generate a trajectory conditioned on time but also on a feature that is time-independent.

Our conditional RNN model  uses static conditional variables such as the length of the trip or number of stops.

\begin{figure}[H]
\centering
\begin{tabular}{cp{0.45\linewidth}}
\includegraphics[width=0.45\linewidth]{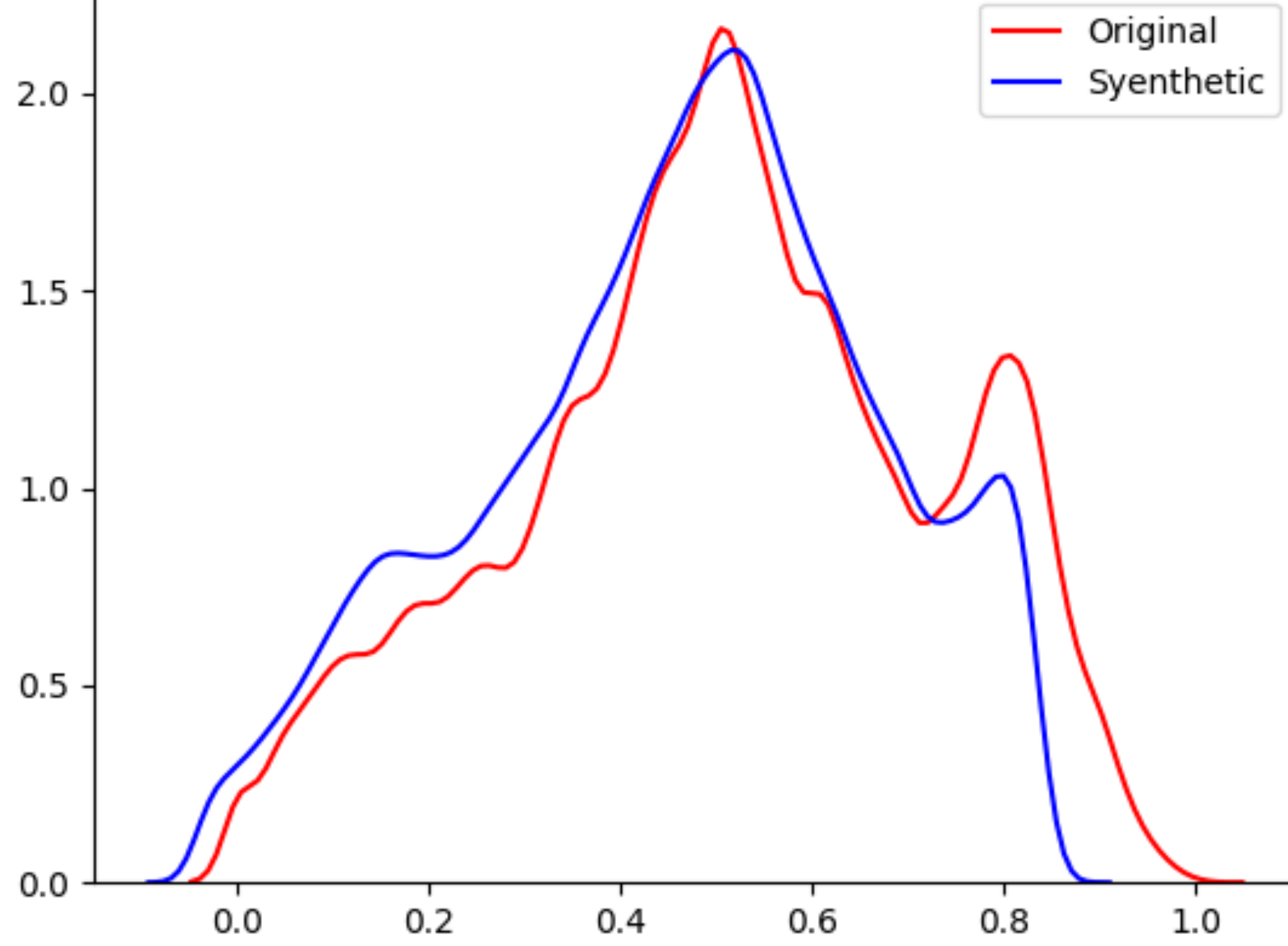} & \includegraphics[width=\linewidth]{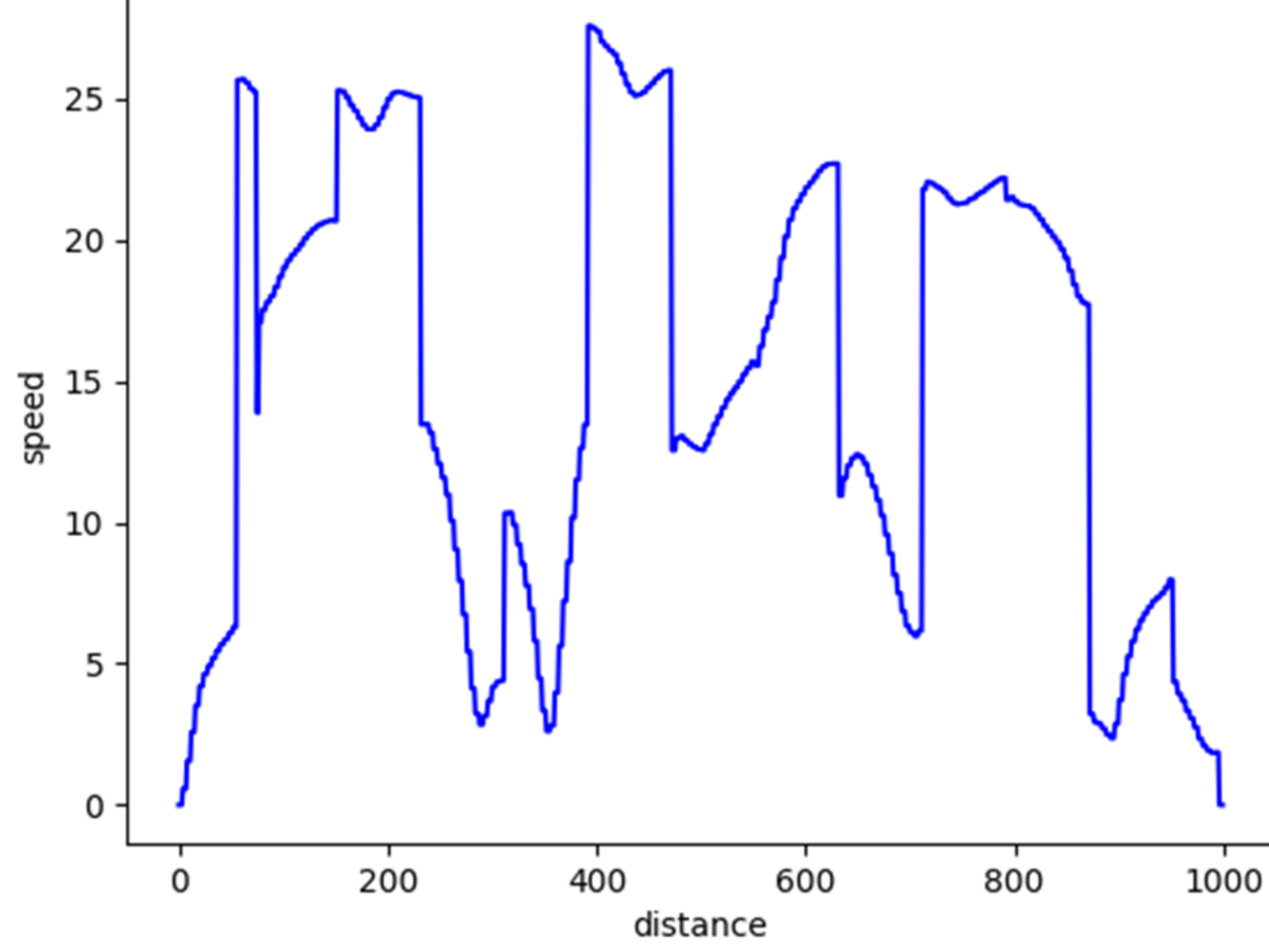}\\
(a) Speed density comparison & (b) Example of a generated speed trajectory
\end{tabular}
\caption{Conditional RNN}
\label{fig:crnn}
\end{figure}

In this section we demonstrate the performance of this model when trip length is used as a conditional variable. As we discussed in Section \ref{sec:method}, the input of both the discriminator and generator have an additional feature that acts as the constraint. At the generation stage, the constraint is also added to the noise vector. Figure \ref{fig:crnn}(a) shows the density plot of conditional model. The trajectories are generated under different length (measured in meters) constraints between 1000 and 6000, and for each constraint 10 different trajectories are generated. Figure \ref{fig:crnn}(b) shows a trajectory generated by the model. As expected, the trajectory starts and ends in zero.

We demonstrated the results for four generative model for trajectory generation, the first model is a combination of $K$ normalizing flows  models. As we can see in Figure \ref{fig:NF_results}, the density distribution of the generated trajectories does not match that of the original training data, and we were not able to identify a normalizing flows model that generates realistic trajectories.  We implemented  three variations of the hybrid AE/GAN generator. The one dimensional RNN generator only takes the speeds as the input, so it can be described as a univariate time series model. The three dimensional generator takes the distance and trip length also as input, so it is a multi-variate time series model.  These two models are used to generate speed trajectories, and as we can see from figures \ref{fig:rnn1d}(a) and \ref{fig:3drnn}(a), the density plots of the three dimensional generator illustrates a more promising model. The Conditional RNN generator enforces a constraint on trajectory generation, so it treats the trip length as a constraint that the generated trajectory has to meet. 

Based on the different generators presented in this section, the hybrid AE/GAN models work best on the speed profiles. If we do not have any conditions imposed on the generation, we can use the 1D-RNN or the 3D-RNN generators, however, the 3D-RNN generator works better because  some additional information implicitly is fed to it as an input. If we need to impose some constraints on the generation process, we can use the C-RNN models to accommodate the additional requirements.

\section{Discussion}\label{sec:conclusion}
Generating realistic  vehicle trajectories is challenging, yet an important sub-problem in several intelligent vehicle problems, such as estimation of energy consumption under different assumptions about routes and powertrains used for for those trips. Our generative neural network, which is a hybrid model consisted of a LSTM-based generative adversarial networks, and a LSTM-based autoencdoer was shown to be a viable alternative to traditional Markov Chain techniques. We use random search to tune hyper-parameters of our model, and trained both generator and discriminator simultaneously.  By deploying hybrid LSTM-based AE/GAN, the trained model is capable of generating trajectories that are close in density distribution to the training data, and can be used as a speed profile in energy optimization problems.  We were able to generate realistic speed profiles that satisfy the constraints that are imposed by the vehicle type and route attributes.  that are  for the Chicago area.  

In this paper, we focused on speed profiles to generate trajectories. However, there are other factors that play a role in speed profiles, such as road and vehicle characteristics. Road characteristics, such as speed limits and stop signs, definitely impact the speed profiles, as well as vehicle characteristics such as engine power and how fast it can accelerate and brake. In our future work, we also take into accounts these characteristics to generate more realistic speed profiles. Another potentially fruitful approach to model the stochastic nature of the speed trajectories is to use neural network with stochastic parameters \cite{polson2017deep}.

\section*{Acknowledgment}

The authors would like to thank Department of Energy and Argonne National Laboratory who sponsored this research.

\ifCLASSOPTIONcaptionsoff
  \newpage
\fi



\renewcommand{\IEEEbibitemsep}{0pt plus 1.0pt}
\makeatletter


\bibliographystyle{IEEEtran}
\bibliography{IEEEabrv,Bibliography}

%




\end{document}